\documentclass[letterpaper]{article} 
\usepackage[submission]{aaai23}  
\usepackage{times}  
\usepackage{helvet}  
\usepackage{courier}  
\usepackage[hyphens]{url}  
\usepackage{graphicx} 
\usepackage{natbib}  
\usepackage{caption} 
\usepackage{algorithm}
\usepackage{algorithmic}
\usepackage{amsmath,amssymb}
\usepackage{graphicx}
\usepackage{tikz}
\usepackage{comment}
\usepackage{color}
\usepackage[accsupp]{axessibility}
\usepackage{bbm}
\usepackage{xcolor}
\usepackage{booktabs}
\usepackage[normalem]{ulem}
\usepackage{multirow}
\usepackage{array}
\usepackage{subfig}
\usepackage{tabularx}
\usepackage{xspace}
\usepackage{newfloat}
\usepackage{listings}
\DeclareCaptionStyle{ruled}{labelfont=normalfont,labelsep=colon,strut=off} 
\floatstyle{ruled}
\newfloat{listing}{tb}{lst}{}
\floatname{listing}{Listing}
\usepackage[capitalize]{cleveref}
\crefname{section}{Sec.}{Secs.}
\Crefname{section}{Section}{Sections}
\Crefname{table}{Table}{Tables}
\crefname{table}{Tab.}{Tabs.}

\title{Contrastive Classification and Representation Learning with Probabilistic Interpretation}

\author {
    Rahaf Aljundi,\textsuperscript{\rm 1}
    Yash Patel, \textsuperscript{\rm 2}
    Milan Sulc \textsuperscript{\rm 2}
    Daniel Olmeda,\textsuperscript{\rm 1}
    Nikolay Chumerin\textsuperscript{\rm 1}\\
      \textsuperscript{\rm 1}Toyota Motor Europe 
\textsuperscript{\rm 2}Visual Recognition Group, Czech Technical University in Prague 
}
  
\DeclareMathOperator{\similarity}{sim}
\newcommand{\z}{\mathbf{z}}
\newcommand{\x}{\mathbf{x}}
\newcommand{\SimCLR}{\text{SimCLR}\xspace}
\newcommand{\SupCon}{\text{SupCon}\xspace}
\newcommand{\CE}{\text{CE}\xspace}
\newcommand{\SPCE}{\text{SPCE}\xspace}

\newcommand{\pt}{\text{pt}\xspace}

\renewcommand{\th}{\boldsymbol{\theta}}
\renewcommand{\c}{\mathbf{c}}
\newcommand{\ExtSupCon}{\tt{ESupCon}\xspace}
\newcommand{\ESupCon}{\text{ESupCon}\xspace}
\newcommand{\tSPCE}{\tt{SPCE}\xspace}
\newcommand{\tCE}{\tt{CE}}
\newcommand{\tsupcontt}{\tt{SupCon+Tt}\xspace}
\newcommand{\tsupconce}{\tt{SupCon+CE}\xspace}
\newcommand{\tsupconceN}{\tt{SupCon+CE(n)}\xspace}

\newcommand{\eg}{e.g.\@\xspace}

\newcommand{\etc}{%
    \@ifnextchar{.}%
        {etc}%
        {etc.\@\xspace}%
}
\begin{document}

\maketitle

\begin{abstract}
Cross entropy loss has served as the main objective function for classification-based tasks. Widely deployed for learning neural network classifiers, it shows both effectiveness and a probabilistic interpretation.  Recently, after the success of self supervised contrastive representation learning methods, supervised contrastive methods have been proposed to learn representations and have shown superior and more robust performance, compared to solely training with cross entropy loss. However, cross entropy loss is still needed to train the final classification layer. In this work, we investigate the possibility of learning both the representation and the classifier using one objective function that combines the robustness of contrastive learning and the probabilistic interpretation of  cross entropy loss. First,  we revisit a previously proposed contrastive-based objective function that approximates cross entropy loss and present a simple extension to learn  the classifier jointly. Second, we propose a new version of the supervised contrastive training that learns jointly the parameters of the classifier and the backbone of the network. We empirically show that our proposed objective functions show a significant improvement over the standard cross entropy loss with more training stability and robustness in various challenging settings.
\end{abstract}

\section{Introduction}\label{sec:into}
\begin{figure*}[t]
    \centering
\includegraphics[width=.6\textwidth]{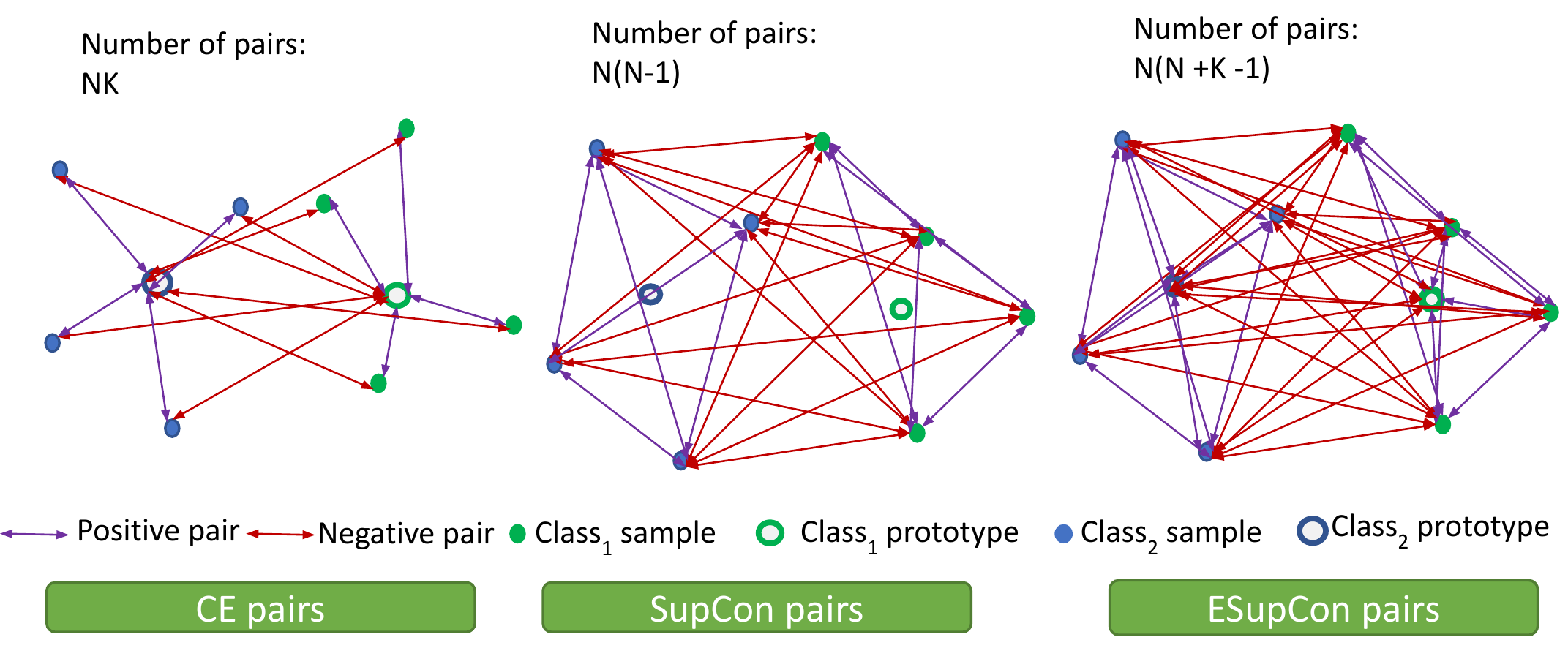}
\caption{\footnotesize Illustration of the possible number of pairs that each loss accesses during training in the learned embedding space, N is the batch size and K is the number of classes.
CE pairs are only defined through classes weights while in SupCon each sample forms positive pairs with its class samples and negative pairs with samples from other classes.
For our ESupCon in addition to the positive and negative samples pairs, class weights (prototypes) form positive pairs with corresponding class samples and negative pairs with other classes samples. Note that here we don't consider augmentations. }%
    \label{fig:illustration}%
\end{figure*}
Representation learning is a powerful tool to create an embedding space that is beneficial for performing downstream tasks e.g., classification or retrieval.  Contrastive representation learning first proposed by ~\cite{chopra2005learning} is a dominant successful line for representation learning. It divides the data into pairs of positive (similar)  and negative (unrelated) samples with the objective of maximizing the similarity of positive pairs samples and minimize it for negative pairs.   


More recently, contrastive learning has become a key component of methods for self-supervised learning~\cite{chen2020simple,kalantidis2020hard,chen2020big,caron2021emerging} and has shown impressive performance~\cite{caron2020unsupervised,caron2021emerging} that is very close to the supervised learning counterpart with cross entropy loss.
Moreover, it was  shown that supervised contrastive  learning marginally outperforms  the \textit{cross entropy loss} in fully supervised image classification~\cite{khosla2020supervised}. Not only for standard supervised classification but it has been applied in continual learning~\cite{davari2022probing}, Out of Distribution Detection~\cite{winkens2020contrastive}, Domain Adaptation~\cite{chen2022contrastive} and many more  showing superior performance to  cross entropy based counterpart. 

Minimizing \textit{cross entropy} (CE) loss is widely used in training deep neural network classifiers,
derived as the maximum likelihood estimate (MLE) of classifier's parameters $\th$ to approximate posterior probabilities $\hat p(\text{class}|\text{observation})$.
~\cite{boudiaf2020unifying} draw the connection among popular pairwise-distance losses and the cross entropy loss, showing that all of them are related to maximizing the mutual information (MI) between the learned embeddings and the corresponding samples' labels.

We emphasize the advantages of the probabilistic interpretation of the CE loss in classification problems.
Such explicit probabilistic interpretation is missing within the embedding spaces trained by popular contrastive learning methods. The posterior estimates $\hat p(\text{class}|\text{observation})$ can be utilized when combining classifiers~\cite{kittler1998combining,breiman1996bagging,ju2018relative}, in adaptation to prior shift~\cite{saerens2002adjusting,sulc2019improving,alexandari2020maximum,sipka2021hitchhiker}, in knowledge distillation~\cite{hinton2015distilling};  out-of-distribution detection~\cite{hendrycks2016baseline} and in many other problems.

In this work, we suggest that one possible reason for the improved performance of supervised contrastive learning  is the inherent access to a large number of samples pairs,  
while the ``pairs'' within the softmax \CE loss are centered around the linear classifier weights. Here we draw an analogy with proxy based loss and consider the linear classifier weights optimized  in the softmax \CE loss as proxies for learning the samples representations.  Proxy base training that utilizes proxies instead of the direct sample to sample relationship is simple and faster to converge, however,  it doesn't leverage the rich data to data similarities as the supervised contrastive loss. 
We refer to Figure~\ref{fig:illustration} for an illustration on this assumption.
We hypothesize that the access to more pairs during training might lead to a better convergence and less overfitting resulting in the advantages hinted in recent works~\cite{khosla2020supervised,graf2021dissecting}. 
 
Hence, to combine the advantages of contrastive representation learning via pairwise losses and the clear probabilistic interpretation of classifiers trained by cross entropy minimization, we present the following  contributions:
First, we consider the weights of the last linear classification layer as prototypes of each class. 
We show that adding a simple  term corresponding to maximizing the similarity between the prototypes and their class samples, leads to an assignment of the prototypes to the mean of each class samples with momentum updates of representation.
This is optimized during the representation training with a supervised contrastive loss~\cite{khosla2020supervised}, resulting in a nearest prototype classifier~\cite{wohlhart2013optimizing}.
Second, we propose an extension to the \textit{supervised contrastive loss}~(\SupCon)~\cite{khosla2020supervised},
where samples of a given class form positive pairs with their class prototype and other classes samples correspond to negative pairs.

We show that the resulting objective   combines in its formulation  the \SupCon loss~\cite{khosla2020supervised} and the standard \CE loss on prototypes related pairs,  preserving the probabilistic interpretation of the predictions. We refer to this loss as \ESupCon (short for \textit{Extended Supervised Contrastive loss}). 

Third, we revisit the Simplified Pairwise Cross Entropy (\SPCE) loss, proposed in the theoretical analysis of~\cite{boudiaf2020unifying}, and compare it with standard \CE loss and  the supervised contrastive learning loss in an extensive experimental evaluation. 

In our experimental evaluation, we not only consider the fully supervised setting but also for the first time a number of challenging settings (low sample regime, imbalanced data and noisy labels). To the best of our knowledge, this is the first comprehensive evaluation of \SupCon loss and the standard  \CE loss in addition to our proposed extensions.
 
We show \ESupCon is  more powerful as a training objective than the standard \CE loss while maintaining a probabilistic interpretation.
and is more robust in challenging and low sample settings. 
Surprisingly, our simple  prototypes similarity term  is more robust than \CE loss for learning a linear classifier after \SupCon in most of the imbalanced and noisy data experiments. 

In the following, we  describe the closely Related Work, then provide a short Background on pairwise losses and the link to Cross Entropy loss, followed by our extension to Supervised Contrastive Loss.
We validate and compare different studied losses in the Experiments, and summarize  our contributions and limitations in Conclusion.

\section{Related Work}\label{sec:relatedwork}
CE loss is a standard and powerful training objective to optimize deep neural networks for classification-related problems. 
For long, the CE loss was believed to be more effective than representation learning losses \eg, metric learning based losses.
For example,~\cite{boudiaf2020unifying} studied the relation of \CE loss to contrastive metric learning losses and showed that the \CE loss also has a contrastive and a tightness part.
The authors suggested that \CE ``does it all'' and that it is easier to optimize compared to its contrastive-learning counterparts.
Recently, self-supervised learning losses have shown great success~\cite{chen2021exploring,grill2020bootstrap,chen2020simple, he2020momentum,chen2020big,grill2020bootstrap,caron2020unsupervised,caron2021emerging} as pretraining methods with only a small performance gap to that of fully supervised learning. The core of the self-supervised methods is the use of rich data augmentation methods to construct positive pairs corresponding to augmented version of a given sample. 
Closer to our work, \cite{li2020prototypical,caron2020unsupervised}  construct clusters and establish cluster assignments through prototypes while learning the embedding space. 
It remains unclear how these losses can be extended to the supervised setting as in our case. 

Inspired by the self-supervised \SimCLR loss~\cite{chen2020simple}, \cite{khosla2020supervised} introduced a new supervised contrastive learning method called \SupCon, which achieved superior results compared to the standard \CE minimization, and which has been shown to be more generalizable and robust to noise. However, the method is only used to train the image representation  and still relies on the \CE loss to train the linear classifier afterwards.
\CE-based training suffers from  known issues of noise sensitive, overfitting~\cite{berrada2018smooth}, and being less transferable than the representation learning counterparts~\cite{chen2020simple}.
Recently, \cite{graf2021dissecting} investigated the difference between the \SupCon~\cite{khosla2020supervised} loss and the \CE loss in the geometry of the targeted representation. 
It was shown that both losses target the same geometric solution, however, \SupCon converges much closer to the target leading to a better generalization performance.

As such, starting from the nice suggested characteristics of the \SupCon loss based training, we propose and study alternatives that can train the whole network (representation and classifier) end-to-end, while preserving both the performance improvements of contrastive representation learning and the clear probabilistic interpretation of the \CE loss. 
We start by considering the classes weights as prototypes for each class samples. We learn these prototypes while maximizing positive pairs similarities and minimizing negative pairs similarities.  Our work hence can be seen as combination of proxy (prototype) based and pairwise based contrastive representation learning. 
Proxy based losses resort to learning a set of  proxies as representative of clusters or classes of samples and optimize the similarities to these proxies rather than the data to data similarities. Proxy NCA~\cite{movshovitz2017no} was the first proxy base metric learning method, it is an approximation of NCA (Neares Component Analysis) using proxies. We note that in the case of learning with class level labels the Proxy NCA matches learning with Softmax Cross Entropy loss when the last classification layer is without a bias term and its weights  are  normalized vectors. Proxy anchor loss~\cite{kim2020proxy}, attempts to combine the benefits of both proxy-based and pairwise losses. While in the main loss formulation only similarities to proxies are considered,     the magnitude of the loss gradient w.r.t. each sample is scaled by the  corresponding proxy similarity proportional to other samples-proxies similarities. 
In general, proxy based losses do not use the proxy at test time and it is unknown how they perform for classification or whether there can exist any probabilistic interpretations. 
Circle loss~\cite{sun2020circle} presents a unified framework for both pairwise and proxy based losses but it adopts an adaptive scaling of the loss depending on how much a given similarity is deviated from its optimum. In doing so, Circle loss abandons the probabilistic interpretation of a sample assignment to its prototype (proxy). 

\section{Background}\label{sec:background}
In this section, we describe recent self-supervised and supervised contrastive losses and the connection with CE loss.
\subsection{Pairwise Losses}
Contrastive losses work with pairs of  embeddings that are pulled together if a pair is positive (related embeddings) and pulled further apart otherwise~\cite{chopra2005learning}. 
 Consider the following: 1) a random data augmentation module that for each sample $\x$ generates two differently augmented samples, 2) a neural network encoder $f$ that maps an augmented input sample $\x$ to its  feature representation: $f(\x)=\z, \z\in \mathbb{R}^d$.  We  start by outlining SimCLR~\cite{chen2020simple}, a popular, effective and simple self supervised contrastive loss to lay the ground for our work:
\begin{equation}
\ell_{\SimCLR}=\frac{1}{2N}\sum_i^N \ell_{\SimCLR}(\z_i,\z_{i+N}) +\ell_{\SimCLR}(\z_{i+N},\z_i),
\end{equation}
\begin{equation}
\begin{split}
\ell_{\SimCLR}(\z_i,\z_j) &= -\log\frac{\exp (\similarity(\z_i,\z_j)/\tau)}{\sum_{k\ne i} \exp(\similarity(\z_i,\z_k)/\tau)},
\end{split}
\end{equation}
where $\tau$ is the temperature scaling term, $N$ is the mini batch size, and the pairs $(\z_i,\z_j)$ consist of features of two differently augmented views of the same data example and $\similarity(\z_i,\z_j) = \dfrac{\z_i^\top \z_j}{||\z_i||\cdot||\z_j|| }$ is the cosine similarity.
Assuming normalized embedding vectors $\z_i$, this pairwise loss is:
\begin{equation}
\ell_{\SimCLR}(\z_i,\z_j) = -\z_i^\top\z_j/\tau + \log \sum_{k\ne i}\exp\left(\z_i^\top\z_k/\tau\right).
\end{equation}
Note that the first term corresponding to the positive pair is the tightness term  and the second one is the contrastive term.
The aforementioned self-supervised batch contrastive approach was extended in~\cite{khosla2020supervised} to the fully supervised setting with the Supervised Contrastive Loss:
\begin{equation}
\ell_\SupCon=\frac{1}{2N}\sum_i^{2N}\ell_\SupCon(\z_i, P_i),
\end{equation}
\begin{equation}\label{eq:supcon}
\begin{split}
&\ell_\SupCon(\z_i, P_i)=\\
&-\frac{1}{|P_i|}\sum_{\z_p\in P_i} \log\frac{\exp(\similarity(\z_i,\z_p)/\tau)}{\sum_{j\ne i}\exp(\similarity(\z_i,\z_j)/\tau)} =\\
&\frac{1}{|P_i|}\sum_{\z_p\in P_i} \left(-{(\z_i^\top\z_p)}{/\tau} + \log \sum_{j\ne i}\exp\left(({\z_i^\top\z_j)}{/\tau}\right)\right),
\end{split}
\end{equation}
where $P_i$ is the set of representations $\z_p$ forming positive pairs for the $i$-th sample, and the index $j$ iterates over all (original and augmented) samples. SupCon loss is expressed as the average  of the loss defined on each positive pair where in this supervised setting, the positive pairs are formed of augmented views and other samples of the same class. 
The authors showed that the supervised contrastive learning achieves excellent results in image classification, improving ImageNet classification accuracy with ResNet-50 by 0.5\% compared to the best results achieved by training with the \CE loss.



\subsection{Cross Entropy and Pairwise Cross Entropy}

The cross entropy~(\CE) loss is a common choice for training classifiers, as its minimization leads to the maximum likelihood estimate of the classifier parameters for estimating the posterior probabilities $\hat p(\text{class} | \text{observation})$.

For $N$ samples of $K$ classes, and a single-label softmax classifier, the \CE loss can be defined as follows:
\begin{equation}
\begin{split}
\ell_\CE &= \frac{1}{N}\sum\limits_{i=1}^N \ell_\CE(\z_i) = -\frac{1}{N}\sum\limits_{i=1}^N\log  \dfrac{\exp{\th_{y_i}^\top\z_i}}{\sum\limits_{k=1}^K \exp{\th_k^\top \z_i} }\\
&= -\frac{1}{N}\sum\limits_{i=1}^N \th_{y_i}^\top \z_i + \frac{1}{N}\sum\limits_{i=1}^N \log \sum\limits_{k=1}^K \exp{\th_k^\top \z_i},
\end{split}
\label{eq:l_CE}
\end{equation}
where $\z_i$ is sample feature for the $i$-th observation having label $y_i\in\{1,\dots,K\}$, and $\th=(\th_1, \dots,\th_K)$ are the parameters of the last fully connected  layer, assuming that no bias term is used.

The \textit{Simplified Pairwise Cross Entropy} (SPCE) loss was introduced  in~\cite{boudiaf2020unifying} as a variant of the \CE loss~\eqref{eq:l_CE}:
\begin{equation}
\begin{split}
\ell_\SPCE &=
-\frac1N\sum_{i=1}^N \log\dfrac{\exp\left(\frac1N\sum_{j:y_j=y_i}\z_j^\top\z_i\right)}{\sum\limits_{k=1}^K \exp\left(\frac1N\sum_{j:y_j=k}\z_j^\top\z_i\right)}.
\end{split}
\label{eq:spce}
\end{equation}
When training the feature encoder with the $\ell_\SPCE$ loss, the classifier weights $\th$ can be estimated directly from the class feature means $\c_k$.
Moreover, the class posterior probabilities $p(k|\z_i)$ also can be estimated explicitly:
\begin{equation}
p(k|\z_i) = 
    \dfrac{\exp\left(\frac{1}{N} \sum_{j:y_j=k}\z_j^\top\z_i\right) }{\sum\limits_{c=1}^K \exp\left( \frac{1}{N}\sum_{j:y_j=c}\z_j^\top\z_i \right)}.
\end{equation}

In the experimental section, we will evaluate \SPCE loss and compare it with \SupCon.
Differently from \SPCE, with \SupCon, one needs to train a classifier on top of the learned representation as a posthoc process. In the following we will discuss and propose alternatives to  jointly learn the classifier  and the feature extraction parameters.

\section{Learning a Classifier Jointly with Representation Learning}\label{sec:jointlearning}
Representation learning under  \SupCon or \SPCE losses targets  grouping one class samples together while pushing  away samples of other classes. In fact, both losses contain  tightness and contrastive terms and  fulfill similar objectives to  that of the cross entropy loss.

Assuming that forcing samples of different classes to lie far apart  is achieved by the contrastive part of \SupCon or \SPCE, in order to learn the parameters of the classifier, one can consider the weight vectors of the linear classifier as prototypes and optimize these prototypes to be closest to the samples of the class they represent (with solely a tightness term).
We assume that both the samples representations  and the classifier weights are normalized vectors and that the classifier is linear with no bias term. We define the following loss to learn the desired prototypes:
\begin{equation}
    \ell_\text{tt}=\frac{1}{N}\sum_i^{N}\ell_\text{tt}(\z_i,\th_{y_i})=\frac{1}{N} \sum_i^{N} -\z_i^\top\th_{y_i}.
    \label{eq:tt}
\end{equation}
Note that the number of samples in~\eqref{eq:tt} might differ from $N$ (\eg, due to augmentation), in which case  $N$  should be replaced by the corresponding number of samples.
With that assumption, the classifier we use is a nearest prototype classifier i.e., assigning a test sample to the class of the nearest prototype.
Note that $\ell_\text{tt}$  resembles only the tightness part of the \CE loss~\eqref{eq:l_CE}. 
The gradient of the $\ell_\text{tt}$ loss w.r.t.\@ the classifier weights can be directly derived from~\eqref{eq:tt}:
\begin{equation}
    \frac{\partial\ell_\text{tt}}{\partial\th_k}=
    -\frac{1}{N} \sum_{i: y_i=k}\z_i.
\end{equation}
Through minimizing this loss jointly with the representation learning loss, we update the classifier weights using the following iterative formula:
\begin{equation}
    \th^{0}_k=\eta\frac{1}{N}\sum_{i: y_i=k}\z_i^0,\;\;\; 
    \th^{t+1}_k=\th^t_k +\eta\frac{1}{N}\sum_{i: y_i=k}\z_i^{t+1},
\end{equation}
where $t$ is the iteration index and $\eta$ is the learning rate.
Note that this is equivalent to setting (up to a constant) the class weights $\th_k$ to the  class features mean $\c_k$ with momentum updates, where the new prototype combines the new iteration representation mean with the previous iteration mean. 
We will compare the minimization of the $\ell_\text{tt}$ loss jointly with the the representation learning loss vs.\@ simply setting the classifier weights $\th_k$ to the hard mean $\c_k$ for each class $k$.

\section{Extended Supervised Contrastive Learning}\label{sec:ESupcon}
Here we aim at extending the \SupCon loss to include the classes prototypes being learned.
For this, we propose to consider an explicit linear classification layer with parameters $\th=(\th_1, \dots,\th_K)$ in the optimization of the supervised contrastive loss~\eqref{eq:supcon}.
Note that here we consider the embeddings $\z_i$ and the class prototypes $\th_k$ in the same feature space.
A class prototype $\th_k$ should  represent as closely as possible its class features. Hence  a  prototype similarity  with its class features should be maximized  and minimized with other classes features. To achieve this we propose to construct the following prototype-feature pair  $(\z_i,\th_{y_i})$ with sample representation $\z_i$ $(y_i=k)$ as a positive pair. 
Now we define the following loss on a positive prototype-feature pair: 
\begin{equation}
  \begin{aligned}[b]
&\ell_{\pt}(\z_i,\th_{y_i}) = -\z_i^\top\th_{y_i}\\ 
&+ \log\left(\sum_{k=1}^{K}\exp(\z_i^\top\th_k)+\sum_{j=1:j\ne i}^{2N} \exp(\z_i^\top\z_j)\right).
\end{aligned}
\label{eq:l_clSup_pairwise}
\end{equation}
Note that SupCon loss on a positive pair of samples is defined as follows:
\begin{equation}
\ell_\SupCon(\z_i,\z_p)=-\z_i^\top\z_p + \log \sum_{j\ne i}\exp({\z_i^\top\z_j)}.
\label{eq:l_SupCon_pairwise}
\end{equation}

Here we omit the temperature $\tau$ for clarity and for a better connection to the \CE loss.
In \eqref{eq:l_clSup_pairwise} we have extended the  set of existing data representations $\z_i$ with the class prototypes $\th_l$. 
Following the same analogy and constructing all positive prototype-feature pairs, the prototype loss  for a class weight $\th_k$ will be defined as follows.
\begin{equation}
\ell_{\pt}(\th_k)=\dfrac{1}{2N_k}\sum_{i:y_i=k} \ell_{\pt}(\z_i,\th_{k}).
    \label{eq:l_clSup}
\end{equation}
Note that the number of summation terms in~\eqref{eq:l_clSup} is $2N_k$ (where $N_k$ is the number of the non-augmented samples in $k$-th class), since the samples in \SupCon are considered with their augmentations.
Having the loss defined per prototype $\th_k$, we can define the full objective function that optimizes the encoder (representation backbone) parameters jointly with the classifier parameters $\th$ as:
\begin{equation}\label{eq:Esupcon}
\ell_{\ESupCon}=\frac{1}{2N+K}\left(\sum_{k=1}^{K}  \ell_\pt(\th_k)  +\sum_{i}^{2N}\ell_\SupCon(\z_i,P_i)\right).
\end{equation}
Next we show that our proposed prototype loss $\ell_{\pt}(\z_i,\th_{k})$ for a given positive pair can be expressed in terms of \SupCon loss on that positive pair and \CE loss  on the concerned sample . 
Let us define the following: 
\begin{equation}
  \begin{aligned}[b]
  T&=\z_i^\top\th_{y_i},\\
  C_1&=\sum_{k=1}^{K} \exp(\z_i^\top\th_k),\\
  C_2&=\sum_{j=1:{j\ne i}}^{2N} \exp(\z_i^\top\z_j),\\
  \exp(\ell_\CE(\z_i)) &= \exp(-T+\log(C_1))\\
        &=\exp(-T)C_1,\\
  \exp(\ell_\SupCon(\z_i, \th_{y_i})) &= \exp\left(-T+\log(C_2 +\exp(T)\right)\\
        &=\exp(-T)(C_2+\exp(T)),
  \end{aligned}
\end{equation}
where $T$ is the tightness term, $C_1$ is the first contrastive term and $C_2$ is the second contrastive term, $\ell_\CE(\z_i)$ is the CE loss for a sample $\z_i$, and 
the SupCon loss $\ell_\SupCon(\z_i,\th_{y_i})$ is estimated after including $\th_{y_i}$ into the pool of representations.

Then our loss for the $(\z_i,\th_{y_i})$ pair can be expressed as:
\begin{equation}
\begin{split}
&\ell_{\pt}(\z_i,\th_{y_i}) = -T + \log(C_1 + C_2)\\
&=\log\left(\exp(-T +\log(C_1 + C_2))\right)\\
&=\log\left(\exp(-T)(C_1 + C_2)\right)\\
&=\log\left(\exp(-T) (C_1 + C_2 + \exp(T) -\exp(T))\right) \\
&=\log(\exp(-T)C_1 + \exp(-T)(C_2 + \exp(T))\\
&-\exp(-T)\exp(T))\\
&=\log\left(\exp(\ell_\CE(\z_i))+  \exp(\ell_\SupCon(\z_i,\th_{y_i}))-1\right).
\end{split}
\end{equation}
%
%
As such, minimizing $\ell_{\pt}(\z_i,\th_{y_i})$ is minimizing the log sum  exponential (LSE) of cross entropy loss and supervised contrastive loss for a given positive pair $(\z_i,\th_{y_i})$, a smooth approximation to the max function.
Note that $\ell_{\pt}(\z_i,\th_{y_i})=0\;\iff\; \ell_{\CE}(\z_i)=\ell_{\SupCon}(\z_i,\th_{y_i})=0$.

We refer to the loss in \eqref{eq:Esupcon} as \ESupCon, short for Extended Supervised Contrastive learning.
In the following, we will extensively compare the different studied loss functions. 

\section{Experiments}\label{sec:exp}
\begin{table}[t]
    \begin{center}
    \footnotesize
    \resizebox{0.5\textwidth}{!}{
        \begin{tabular}{l|l|l|l|l|l}
    \toprule
    Method &
    CIFAR-10 &
    CIFAR-100 &
    Tiny ImageNet &
     Caltech256 &
    Avg.
    \\

        \midrule
    {\tCE}  &
    $95.39$ & %
    $76.36$ &
    $65.76$&
    $55.9$\ &
{$-$}
    \\
    \midrule
    *{\tsupconce}  & 
    $95.50$ \textcolor{blue}{$+0.11$} & 
    $75.90$ \textcolor{red}{$-0.46$} &
    $65.56$ \textcolor{red}{$-0.20$} &
    $57.91$ \textcolor{blue}{$+2.01$} &
    \textcolor{blue}{$+0.36$}
    \\
    
    *{\tsupconceN} &
    $95.27$ \textcolor{red}{$-0.12$} & 
    $74.57$ \textcolor{red}{$-1.79$} &
    $61.69$ \textcolor{red}{$-4.07$} &
     $52.92$ \textcolor{red}{$-2.98$} &
    \textcolor{red}{$-1.52$}
    \\
    
    *{\tsupcontt} &
    $95.20$ \textcolor{red}{$-0.19$} &
    $74.80$ \textcolor{red}{$-1.56$} & 
    $59.66$ \textcolor{red}{$-6.1$} &
    $57.42$ \textcolor{blue}{$1.52$} &
    \textcolor{red}{$-2.24$} 
    \\
    \midrule
    
    {\tSPCE} & 
    ${95.62}$ \textcolor{blue}{$+0.23$} &
    $\underline{78.15}$ \textcolor{blue}{$+ 1.79$} &
    $\underline{66.52}$ \textcolor{blue}{$+0.76$} &
     ${48.46}$ \textcolor{red}{$-7.44$} &
    \textcolor{red}{$-1.16$}
    \\
    
    {\tSPCE}({\tt M}) &
    $95.30$ \textcolor{red}{$-0.09$} & 
    $77.49$ \textcolor{red}{$+1.13$} &
    $66.28$ \textcolor{blue}{$+0.52$} &
     $48.37$ \textcolor{red}{$-7.52$}&
    \textcolor{red}{$-1.49$}
    \\

    {\ExtSupCon} & 
   $\underline{ 95.9}$ \textcolor{blue}{$+0.51$} & 
    $76.92$ \textcolor{blue}{$+0.56$} & 
    $66.2$ \textcolor{blue}{$+0.44$} &
     $\underline{58.27}$ \textcolor{blue}{$+2.37$} &
    \textcolor{blue}{$\underline{+0.97}$}
    \\
    \bottomrule
    \end{tabular}}
    \caption{\footnotesize Accuracy $(\%)$ of the different studied and proposed losses on fully labelled  datasets. * indicates the use of a projection head. Absolute gains over cross entropy are reported \textcolor{blue}{blue} and absolute declines in \textcolor{red}{red}. The last column shows an average improvement or decline over {\tCE}, across the datasets.}
    \label{tab:full_dataset}
    \end{center}
    \vspace{-2em}
\end{table}

This section serves to compare the performance of deep models trained under the different objective functions discussed earlier including tightness loss term~\eqref{eq:tt} and \ESupCon~\eqref{eq:Esupcon}. Our goal is to perform an extensive evaluation of the different losses behaviour not only under fully supervised setting but also under more challenging yet more plausible settings, namely limited data, imbalanced data and noisy labels settings. For the purpose of this experimental validation,  we focus on the object recognition problem.  

\subsection{Datasets}
We consider Cifar-100, Cifar-10~\cite{krizhevsky2009learning}, Tiny ImageNet~\cite{tinyimgnet} (a subset of $200$ classes from ImageNet~\cite{deng2009imagenet}, rescaled to the $32\times32$) datasets and Caltech256~\cite{griffin2007caltech}.
We refer to the supplementary materials for more results.

\begin{table*}[t]
  \vspace{-2em}
    \begin{center}
    \footnotesize
    \setlength{\tabcolsep}{2pt}
    \resizebox{\textwidth}{!}{
        \begin{tabular}{l|l|l|l|l|l|l|l|l|l|l}
    \toprule
    \multirow{2}[1]{*}{Method} &
    \multicolumn{3}{c|}{CIFAR-10} &
    \multicolumn{3}{c|}{CIFAR-100} &
    \multicolumn{3}{c|}{Tiny ImageNet} &
    \multirow{2}[1]{*}{Avg.}
    \\
    \cline{2-10}
     & 
     $N=2\text{K}$ &
     $N=5\text{K}$ &
     $N=10\text{K}$ &
     $N=8\text{K}$ &
     $N=10\text{K}$ &
     $N=20\text{K}$ &
     $N=20\text{K}$ &
     $N=50\text{K}$ &
     $N=70\text{K}$ &
     \\
    \midrule 

    \midrule
        {\tCE} &
    $28.02$ &
    $69.91$&
    $85.08$ &
    $43.67$ &
    $51.09$&
    $64.31$&
    $44.29$ &
    $57.19$&
    $60.94$ &
   -
    \\
    \midrule
    *{\tsupconce}&
    $72.27$ \textcolor{blue}{$+44.25$} & 
    $82.37$ \textcolor{blue}{$+12.46$} &
    $88.03$ \textcolor{blue}{$+2.95$} &
    $50.96$ \textcolor{blue}{$+7.2$} &
    $54.49$ \textcolor{blue}{$+3.4$} &
    $64.39$ \textcolor{blue}{$+ 0.08$} &
    $44.00$ \textcolor{blue}{$-0.29$} &
    $59.24$ \textcolor{blue}{$+2.05$} &
    $62.88$ \textcolor{blue}{$+1.94$} &
    \textcolor{blue}{$+8.22$}
    \\

    *{\tsupconceN} &
    $71.99$ \textcolor{blue}{$+43.97$} & 
    $82.73$ \textcolor{blue}{$+12.82$} &
    $87.91$ \textcolor{blue}{$+ 2.83$} &
    $50.60$ \textcolor{blue}{$+6.93$} &
    $53.92$ \textcolor{blue}{$+ 2.83$} &
    $63.27$ \textcolor{red}{$ -1.04$} &
    $43.50$ \textcolor{red}{$-0.79$} &
    $57.62$ \textcolor{blue}{$+0.43$} &
    $59.62$ \textcolor{red}{$-1.32$} &
    \textcolor{blue}{$+7.41$}
    \\

    *{\tsupcontt} &
    $72.17$ \textcolor{blue}{$+44.15$} &
    $82.97$ \textcolor{blue}{$+13.06$} &
    $87.37$ \textcolor{blue}{$+2.29$} &
    $51.23$ \textcolor{blue}{$+7.56$} &
    $54.49$ \textcolor{blue}{$+3.4$} &
    $64.28$ \textcolor{red}{$-0.02$} &
    $43.82$ \textcolor{red}{$-0.47$} &
    $52.21$ \textcolor{red}{$-4.98$} &
    $57.88$ \textcolor{red}{$-3.06$} &
    \textcolor{blue}{$+6.88$}
    \\
    \midrule
    
    {\tSPCE} &

    $31.81$ \textcolor{blue}{$+3.79$} &
    $78.60$ \textcolor{blue}{$+8.69$}&
    $86.15$ \textcolor{blue}{$+1.07$} &
    $50.09$ \textcolor{blue}{$+6.42$} &
    $53.78$ \textcolor{blue}{$+2.69$} &
    $64.82$ \textcolor{blue}{$+0.51$} &
    $40.94$ \textcolor{red}{$-3.35$} &
    $55.70$ \textcolor{red}{$-1.49$} &
    $55.49$ \textcolor{red}{$-5.45$} &
    \textcolor{blue}{$+1.43$}
    \\
    
   {\ExtSupCon} & 

       $74.08$ \textcolor{blue}{$+46.06$} &
    $83.89$ \textcolor{blue}{$+13.98$} &
    $88.83$ \textcolor{blue}{$+3.75$} &
    $48.26$ \textcolor{blue}{$+4.59$} &
    $52.58$ \textcolor{blue}{$+1.49$} &
    $63.12$ \textcolor{red}{$-1.19$} &
    $44.17$ \textcolor{red}{$ -0.12$} &
    $58.66$ \textcolor{blue}{$+1.47$} &
    $62.62$ \textcolor{blue}{$+ 1.68$} &
    \textcolor{blue}{$+7.97$}
   \\
    \bottomrule
    \end{tabular}}
    \caption{\footnotesize Accuracy $(\%)$ on CIFAR-10, CIFAR-100 and Tiny ImageNet for a low-sample training scenario, where $N$ represents the number of samples used for the training. Absolute gains over cross entropy are reported in \textcolor{blue}{blue} and absolute declines in \textcolor{red}{red}. * indicates the use of a projection head. The last column shows an average improvement or decline over cross entropy ({\tCE}), across the datasets and the settings.}
    \label{tab:lowsamples}
    \end{center}
    \vspace{-1em}
\end{table*}
\begin{table*}[t]
    \begin{center}
    \footnotesize
    \setlength{\tabcolsep}{2pt}
    \resizebox{\textwidth}{!}{
        \begin{tabular}{l|l|l|l|l|l|l|l|l|l|l}
    \toprule
    \multirow{2}[1]{*}{Method} &
    \multicolumn{3}{c|}{CIFAR-10} &
    \multicolumn{3}{c|}{CIFAR-100} &
    \multicolumn{3}{c|}{Tiny ImageNet} &
    \multirow{2}[1]{*}{Avg.}
    \\
    \cline{2-10}
     & 
     $\text{IR}=0.05$ &
     $\text{IR}=0.1$ &
     $\text{IR}=0.5$ &
     $\text{IR}=0.05$ &
     $\text{IR}=0.1$ &
     $\text{IR}=0.5$ &
     $\text{IR}=0.05$ &
     $\text{IR}=0.1$ &
     $\text{IR}=0.5$ &
     \\
    \midrule 

    
    {\tCE} &
    $82.85$ &%
    $87.83$&
    $93.99$&
    $48.57$&
    $54.44$&
    $71.19$ &
    $40.65$&
    $46.16$&
    $60.30$ & 
{$-$}
    \\
    \midrule
    *{\tsupconce}&
    $79.94$ \textcolor{red}{$-2.91$} & 
    $86.86$ \textcolor{red}{$-0.97$} &
    $94.34$ \textcolor{blue}{$+0.35$} &
    $46.79$ \textcolor{red}{$-1.78$} &
    $44.21$ \textcolor{red}{$-10.23$} &
    $71.13$ \textcolor{red}{$-0.06$} &
    $44.96$ \textcolor{blue}{$+4.31$} &
    $49.57$ \textcolor{blue}{$+3.41$} &
    $62.45$ \textcolor{blue}{$+2.15$} &
    \textcolor{red}{$-0.64$}
    \\

    *{\tsupconceN} &
    $47.77$ \textcolor{red}{$-35.08$} & 
    $47.64$ \textcolor{red}{$-40.19$} &
    $90.14$ \textcolor{red}{$-3.85$} &
    $40.00$ \textcolor{red}{$-8.57$} &
    $40.32$ \textcolor{red}{$ -14.12$} &
    $55.94$ \textcolor{red}{$-15.25$} &
    $35.69$ \textcolor{red}{$-4.96$} &
    $35.61$ \textcolor{red}{$-10.55$} &
    $37.70$ \textcolor{red}{$-22.60$} &
    \textcolor{red}{$-17.24$}
    \\
    
    *{\tsupcontt} &
    $85.62$ \textcolor{blue}{$+2.7$} &
    $88.76$ \textcolor{blue}{$+0.93$} &
    $94.40$ \textcolor{blue}{$+ 0.41$} &
    $54.40$ \textcolor{blue}{$+5.83$} &
    $56.79$ \textcolor{blue}{$+ 2.35$} &
    $70.38$ \textcolor{red}{$-0.81$} &
    $44.11$ \textcolor{blue}{$+3.46$} &
    $47.39$ \textcolor{blue}{$+1.23$} &
    $57.30$ \textcolor{red}{$ -3.00$} &
    \textcolor{blue}{$+1.46$}
    \\
    \midrule
    
    {\tSPCE} &
    $85.62$ \textcolor{blue}{$+2.7$} &
    $86.94$ \textcolor{red}{$-0.89$}&
    $93.95$ \textcolor{red}{$-0.04$} &
    $49.59$ \textcolor{blue}{$+1.02$}  &
    $53.78$ \textcolor{red}{$-0.66$}  &
    $68.61$ \textcolor{red}{$-2.58$} &
    $37.27$ \textcolor{red}{$-3.38$} &
    $40.55$ \textcolor{red}{$-5.61$} &
    $61.14$ \textcolor{blue}{$0.84$} &
    \textcolor{red}{$-0.95$}
    \\
    
      {\ExtSupCon} & 
   $86.00$ \textcolor{blue}{$+3.15$} &
    $89.26$ \textcolor{blue}{$+1.43$} &
    $94.77$ \textcolor{blue}{$+0.78$} &
    $52.74$ \textcolor{blue}{$+ 4.17$} &
    $58.08$ \textcolor{blue}{$+3.64$} &
    $71.37$ \textcolor{blue}{$+0.18$} &
    $45.55$ \textcolor{blue}{$ +4.90$} &
    $50.90$ \textcolor{blue}{$+4.74$} &
    $63.08$ \textcolor{blue}{$+2.78$} &
    \textcolor{blue}{$+2.86$}
    \\
    \bottomrule
    \end{tabular}}
    \caption{\footnotesize Accuracy $(\%)$ on CIFAR-10, CIFAR-100 and Tiny ImageNet for an imbalanced training scenario, where IR represents the rate of imbalance. Absolute gains over cross entropy are reported \textcolor{blue}{blue} and absolute declines in \textcolor{red}{red}.* indicates the use of a projection head. The last column shows an average improvement or decline over cross entropy ({\tCE}), across the datasets and the settings.}
    \label{tab:imbalancesamples}
    \end{center}
    \vspace{-1em}
\end{table*}

\subsection{Methods and Implementation Details}
In all experiments we use ResNet50 as a main network and
 evaluate the following losses:

{\tCE}: we  optimize the network parameters using the standard CE loss.
For the SupCon loss~\cite{khosla2020supervised}, we use the publicly available implementation, which uses L2-normalized outputs of a multi-layer head (FC, ReLU, FC), a projection head, on top of the embeddings used  for classification. We learn the classifier parameters using:
i) Cross entropy loss ({\tsupconce}), on the linear layer after optimizing minimizing SupCon loss.  ii) For the sake of fair comparison with other losses, we consider also cross entropy loss with no bias term,  normalized embeddings and normalized classifier weights. We denote this variant by {\tsupconceN}.
iii) Tightness loss ({\tsupcontt}), where we optimize the parameters of a linear classifier using \eqref{eq:tt} during the optimization of the rest of the network (projection head + backbone) with SupCon loss.  Note that  the gradients of the tightness loss are not propagated to the rest of the network.
 {\tSPCE}: we optimize the backbone with SPCE loss~\eqref{eq:spce} and  the classifier weights with the tightness term~\eqref{eq:tt}.
We also show the performance with directly assigning the weights to the mean of each class samples {{\tSPCE}({\tt M}) }. 

 Our {\ExtSupCon}: with~\eqref{eq:Esupcon} we optimize jointly a linear classifier and the backbone parameters.

Note that  {{\tsupconce}, {\tsupconceN} and {\tsupcontt}} use a projection head, unlike {\tCE}, {\tSPCE} and {\ExtSupCon}.
All studied variants benefit from the same type of data augmentations and hyper-parameters  were estimated on Cifar-10 dataset and fixed for the rest. We refer to the supplementary materials for more details.

\subsection{Fully Supervised Classification}
We first start by comparing the different studied methods  on the standard classification setting while leveraging all the labelled training data of each dataset. Table~\ref{tab:full_dataset} shows the average test  accuracy at the end of the training on the three considered datasets. 

First,    {\ExtSupCon} outperforms {\tCE} training alone, using the same number of parameters. {\tsupconce} improves over  {\tCE}. {\tsupcontt} is comparable to {\tsupconceN}.

{\ExtSupCon} shows the best performance on all four datasets.
Except from Caltech dataset,  {\tSPCE} achieves superior results to {\tCE}. When assigning the classifier weights directly to the mean of the features, {\tSPCE}({\tt M}), results are slightly inferior to the use of our tightness loss~\eqref{eq:tt} for training the classifier parameters. For the rest the of experiments, we show only {\tSPCE}, using the suggested tightness term to train the classifier parameters.

\subsection{Classification in Low-Sample Scenario}
\begin{table*}[h!]
  \vspace{-2em}
    \begin{center}
    \footnotesize
    \setlength{\tabcolsep}{2pt}
    \resizebox{\textwidth}{!}{
        \begin{tabular}{l|l|l|l|l|l|l|l|l|l|l}
    \toprule
    \multirow{2}[1]{*}{Method} &
    \multicolumn{3}{c|}{CIFAR-10} &
    \multicolumn{3}{c|}{CIFAR-100} &
    \multicolumn{3}{c|}{Tiny ImageNet} &
    \multirow{2}[1]{*}{Avg.}
    \\
    \cline{2-10}
     & 
     $\text{NR}=0.5$ &
     $\text{NR}=0.3$ &
     $\text{NR}=0.2$ &
     $\text{NR}=0.5$ &
     $\text{NR}=0.3$ &
     $\text{NR}=0.2$ &
     $\text{NR}=0.5$ &
     $\text{NR}=0.3$ &
     $\text{NR}=0.2$
     \\
    \midrule 

        {\tCE} &
    $60.88$ &%
    $87.08$ &%
    $88.93$  &%
    $35.47$ &
    $56.57$&
    $64.93$&
    $31.55$ &
    $49.62$ &
    $55.40$ &
-
    \\
    \midrule
    *{\tsupconce}&
    $48.08$  \textcolor{red}{$-12.8$} & 
    $74.47$  \textcolor{red}{$-12.61$} &
    $85.94$  \textcolor{red}{$-2.99$} &
    $34.78$  \textcolor{red}{$-0.69$} &
    $58.06$  \textcolor{blue}{$+1.49$} &
    $65.57$  \textcolor{blue}{$+0.64$} &
    $31.81$  \textcolor{blue}{$+0.26$} &
    $46.20$  \textcolor{blue}{$+3.42$} &
    $54.74$  \textcolor{blue}{$+0.66$} &
    \textcolor{red}{$-3.42$}
    \\

    *{\tsupconceN} &
    $46.35$  \textcolor{red}{$-14.53$} & 
    $77.42$  \textcolor{red}{$-9.66$} &
    $87.45$  \textcolor{red}{$-1.48$} &
    $33.55$  \textcolor{red}{$-1.92$} &
    $62.44$  \textcolor{blue}{$+5.87$} &
    $67.87$  \textcolor{blue}{$+ 2.94$} &
    $28.88$  \textcolor{red}{$-2.67$}&
    $54.64$  \textcolor{blue}{$+5.02$} &
    $58.62$  \textcolor{blue}{$+3.22$} &
    \textcolor{red}{$-1.47$}
    \\
    
    *{\tsupcontt} &
    $58.05$  \textcolor{red}{$-2.83$} &
    $89.70$  \textcolor{blue}{$+2.62$} &
    $90.66$  \textcolor{blue}{$+ 1.73$} &
    $37.23$  \textcolor{blue}{$+1.76$} &
    $67.76$  \textcolor{blue}{$+11.19$} &
    $69.41$  \textcolor{blue}{$+4.48$} &
    $28.67$  \textcolor{red}{$-2.88$} &
    $54.81$  \textcolor{blue}{$+5.19$} &
    $57.93$  \textcolor{blue}{$+2.53$} &
    \textcolor{blue}{$+2.64$}
    \\
    \midrule
    
    {\tSPCE} &
    $65.63$  \textcolor{blue}{$+4.75$} &
    $88.77$  \textcolor{blue}{$+1.69$} &
    $88.93$  \textcolor{blue}{$+0.0$} &
    $36.56$  \textcolor{blue}{$+1.09$} &
    $60.35$  \textcolor{blue}{$+3.78$} &
    $65.75$  \textcolor{blue}{$+0.82$} &
    $25.27$  \textcolor{red}{$-6.28$} &
    $42.45$  \textcolor{red}{$-7.17$} &
    $49.52$  \textcolor{red}{$-5.88$} &
    \textcolor{red}{$-0.80$}
    \\
    
  {\ExtSupCon}& 
      $59.18$ \textcolor{red}{$-1.70$} &
    $88.15$  \textcolor{blue}{$+1.07$} &
    $90.92$  \textcolor{blue}{$+1.99$}&
    $37.04$ \textcolor{blue}{$+1.57$} &
    $62.94$  \textcolor{blue}{$+6.37$} &
    $65.81$  \textcolor{blue}{$+0.87$} &
    $32.555$  \textcolor{blue}{$+1.0$} &
    $52.80$  \textcolor{blue}{$+3.18$} &
    $56.79$  \textcolor{blue}{$+1.39$} &
    \textcolor{blue}{$+1.75$}
    \\
    \bottomrule
    \end{tabular}}
    \caption{Accuracy $(\%)$ on CIFAR-10, CIFAR-100 and Tiny ImageNet for a noisy training scenario,  NR represents the rate of noise. Absolute gains over cross entropy are reported in \textcolor{blue}{blue} and absolute declines in \textcolor{red}{red}. * indicates the use of a projection head. The last column shows an average improvement or decline over cross entropy ({\tCE}), across the datasets and the settings.}
    \label{tab:noisysamples}
    \end{center}
    \vspace{-2em}
\end{table*}
\begin{figure*}[h!]
    \centering
    \subfloat[{\tCE}]{
    \begin{tabular}{c}\hspace*{-5mm}\includegraphics[width=.20\textwidth]{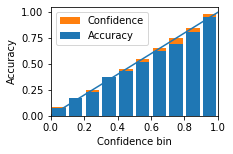}
    \\ $\text{ECE}=0.0300$
    \end{tabular}
    }
    \subfloat[{\tsupconce}]{
    \begin{tabular}{c}\hspace*{-5mm}\includegraphics[width=.20\textwidth]{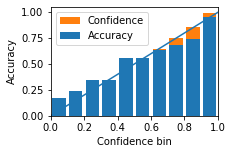}
    \\ $\text{ECE}=0.0492$
    \end{tabular}
    }
    \subfloat[{\tsupcontt}]{
    \begin{tabular}{c}\hspace*{-5mm}\includegraphics[width=.20\textwidth]{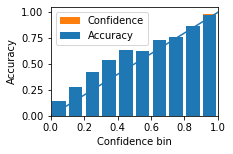}
    \\ $\text{ECE}=0.0621$
    \end{tabular}
    }
    \subfloat[{\ExtSupCon}]{
    \begin{tabular}{c}\hspace*{-5mm}\includegraphics[width=.20\textwidth]{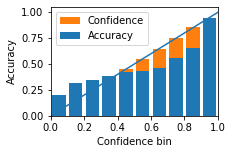}
    \\ $\text{ECE}=0.0606$
    \end{tabular}
    }
    \subfloat[{\tSPCE}({\tt M})]{
        \begin{tabular}{c}\hspace*{-5mm}\includegraphics[width=.20\textwidth]{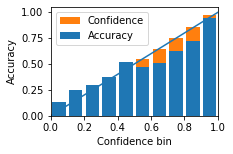}
    \\ $\text{ECE}=0.0562$
    \end{tabular}
    }
    \caption{\footnotesize Reliability Diagrams and Expected Calibration Error of probabilistic classifiers learned with different studied loss functions and further calibrated by temperature scaling.}
    \label{fig:reliability_diagrams}
\end{figure*} 

After studying the fully labelled scenario,  here, we are interested in the performance under limited data setting. Our goal is to see how prone each method is to overfitting in low data regime and whether significant differences can be observed among the different alternatives. Table~\ref{tab:lowsamples} reports the average test accuracy on Cifar-10, Cifar-100 and Tiny ImageNet  using different numbers of training samples ($N$).

While {\tCE} performance is comparable to other losses on the full data scenario, here it is significantly lower than other competitors with a gap increasing as the sample size gets smaller. Except from Tiny ImageNet, {\tsupcontt}  shows comparable performance to {\tsupconce} and is slightly inferior ($0.5\%$) to {\tsupconceN} on average. {\tSPCE} results are  better than {\tCE} on Cifar-10 and Cifar-100.  {\ExtSupCon} improves significantly over {\tCE} while being comparable with {\tsupconce}, however, with no projection head.
{\ExtSupCon} is much more robust than  {\tSPCE} in this setting. 
  
\subsection{Classification under Imbalanced Data}

Our goal is to compare the performance of a model trained by the different studied losses under various challenging settings beside the standard fully supervised setting. 
Here, we examine the scenario where training data are not uniformly distributed.
Some classes are undersampled while others are oversampled. Specifically, we want to test the ability of the different losses to cope with this data nature and  learn  the underrepresented classes.
We simulate this scenario by altering the training data in which half of the categories are underrepresented with a number of samples equals to the imbalance rate (IR) of other categories samples.
The test set on which we report the average accuracy remains balanced.

Table~\ref{tab:imbalancesamples} reports the average test accuracy of models trained to minimize the different losses on the three considered datasets.
For each dataset we consider imbalance rates of $0.05$, $0.1$, and $0.5$ where, for example, an imbalance rate of $0.1$ means that the size of undersampled classes samples is $0.1$ compared to the oversampled classes size.

Here it seems that  {\tsupconce} doesn't improve over {\tCE} alone.    {\tSPCE}   results  are marginally lower than {\tCE}. 
Our two proposed losses {\tsupcontt} and {\ExtSupCon} exhibit more robust and powerful performance compared to {\tCE} with {\ExtSupCon} performing the best. 


\subsection{Classification under Noisy Data}
We continue our investigation on the different losses performance under challenging setting and test 
 another interesting scenario: classification with noisy labels.
We want to test the ability of the different training regimes to  learn generalizable decision boundaries in spite of the presence of wrongly labelled samples. 
To simulate this scenario, during training a percentage of the training data, denoted by noise rate (NR), is associated with wrong labels (shuffled labels).
As in the previous experiments, we report the results on the standard, correctly labelled, test set.
Table~\ref{tab:noisysamples} reports the average test accuracy on Cifar-10, Cifar-100 and Tiny ImageNet with noise rates of ($0.2,0.3,0.5$).
Here we obtained similar results to the imbalanced settings,  {\tsupconce} doesn't consistently improve over {\tCE}, same applies for {\tSPCE}. 
Our both proposed losses improve over  {\tCE} with {\tsupcontt} performing the best here. 
\subsection{General Remarks}
 We note the following on the shown results of the different losses: CE training after {\SupCon} pretraining ({\tsupconce}) improves over standard \CE in full and low data regime. However, deploying {\CE} to learn the classifier with or without  {\SupCon} pretraining is sensitive to noise and data imbalance. Interestingly, our proposed tightness term is more effective on these two scenarios, however inferior on the full and low data regime. 
In all studied settings, our proposed   {\ExtSupCon} loss improves over {\CE}  and over ({\tsupconce}) on the challenging imbalanced and noisy settings.   
In Supplementary we discuss the computational complexity of the different losses and their sensitivity to hyper-parameters.
\subsection{Classifier Outputs as Posterior Probabilities}
To access the interpretation of the classifier outputs as estimates of posterior probabilities $\hat p(\text{class}|\text{observation})$, we calibrated the outputs by temperature scaling~\cite{guo2017calibration} -- we estimated the temperature on a holdout set ($20\%$ of the test set) and computed the reliability diagram and the expected calibration error (ECE) on the remaining test samples of Cifar-100 dataset.
Results are shown in Figure~\ref{fig:reliability_diagrams}: while the standard CE loss has the lowest calibration error, all other calibrated classifiers provide reliable predictions, an interesting result given the shown performance advantage.

\section{Conclusion}\label{sec:conclusion}
In this work, we derive novel, robust objective functions, inspired by new evidence showing that contrastive losses improve performance over \CE. Driven by the question of whether cross entropy loss is the best option to train jointly a good representation and powerful, generalizable, decision boundaries, we start from  a recent approximation to cross entropy loss (\SPCE) with pairwise  training of representation where classifier weights can be assigned to the mean of each class features. We then suggest to learn the classifier weights under only a tightness term jointly with \SupCon representation training or \SPCE. Next, we propose an extension to \SupCon, where the classifier weights are treated as learnable prototypes in the same space as the samples embeddings, and where data points form positive pairs with their classes prototypes. We show that the proposed loss for a given pair $(\z_i, \th_k)$ is a smooth approximation to the maximum of the \CE and \SupCon losses on that pair. To this point, we test the performance of models trained with the different discussed losses under different challenging settings. We show that the proposed extensions demonstrate more robust and stable performance across different settings and datasets. As a future work, we plan to extend the experiments to object detection and image segmentation problems, as well as to test the discussed losses on Out-Of-Distribution and Continual Learning benchmarks.

\bibliography{aaai23.bib}

\end{document}